\documentclass[conference]{IEEEtran}
\IEEEoverridecommandlockouts

\usepackage{cite}
\usepackage{amsmath,amssymb,amsfonts}
\usepackage{graphicx}
\usepackage{booktabs}
\usepackage{textcomp}
\usepackage{xcolor}
\usepackage{url}

\def\BibTeX{{\rm B\kern-.05em{\sc i\kern-.025em b}\kern-.08em
    T\kern-.1667em\lower.7ex\hbox{E}\kern-.125emX}}

\usepackage{eso-pic}
\AddToShipoutPictureBG*{%
  \AtPageLowerLeft{%
    \makebox[\paperwidth]{%
      \raisebox{0.32in}{%
        \parbox{0.84\paperwidth}{\centering\scriptsize
          \copyright~2026 IEEE. Personal use of this material is permitted. Permission from IEEE must be obtained for all other uses, in any current or future media, including reprinting/republishing this material for advertising or promotional purposes, creating new collective works, for resale or redistribution to servers or lists, or reuse of any copyrighted component of this work in other works.\\[2pt]
          Accepted for publication in the 2026 IEEE 8th International Conference on Cognitive Machine Intelligence (CogMI).}}}}}
          
\begin{document}

\title{The Long Road to the Same Answer: Cognitive Bias Under Escalating Reasoning Budgets in Large Language Models}

\author{\IEEEauthorblockN{Obada Kraishan}
\IEEEauthorblockA{College of Media and Communication \\
Texas Tech University \\
Lubbock, TX, USA \\
omareikr@ttu.edu}}

\maketitle

\begin{abstract}
Reasoning models allocate extra computation at inference time and present their answers as the product of deliberate thought. If this deliberation works the way dual-process accounts of human cognition suggest, longer thinking should weaken the classic decision biases that fast, intuitive judgment produces. We test that prediction. Using 30 vignettes covering six biases (anchoring, framing, loss aversion, escalation of commitment, availability, confirmation) drawn from an established benchmark, we run a dose-response study across four model families, pairing each reasoning model with a matched non-reasoning sibling and requesting thinking ceilings of 0, 1,024, 4,096, and 8,192 tokens, for 12,350 API calls. Because a requested ceiling is not the same as realized deliberation, we log the reasoning tokens each call consumed and use realized consumption as the dose. Three results follow. First, reasoning models are not less biased than their siblings; the point estimate leans the other way in every family, but the item-level pooled contrast is not reliable (Delta = +0.031, t(29) = 1.45, p = .157). Second, bias magnitude does not reliably fall as realized deliberation grows: across the ranges these models actually produced, no slope is significantly negative, and where anything moves it is the signed score drifting further from the human direction. Third, anchoring is the only bias in the human direction (d = 1.89). Four of the other five lean the opposite way in all seven models; at the item level, with five items per bias, that reversal is reliable for framing and directional for escalation of commitment, confirmation, and loss aversion, while availability is absent. A one-line instruction to restate the anchor before answering lowered anchoring on all five anchoring items, which no amount of additional thinking did, although with five items the effect does not reach significance (p = .057). The results argue against treating test-time reasoning as a rationality guarantee and for auditing deployed models bias by bias.
\end{abstract}


\section{Introduction}

Reasoning models such as OpenAI's o-series, DeepSeek-R1, and the thinking variants of Claude and Qwen spend a controllable budget of tokens on an internal deliberation phase before committing to an answer \cite{jaech2024openai, guo2025deepseek, yang2025qwen3}. The appeal is intuitive: a system that thinks longer should decide better, and test-time compute has in fact improved performance on mathematics, coding, and other verifiable tasks \cite{snell2024scaling, muennighoff2025s1}. These models are now placed in decision support, where the questions are not math problems but judgment calls: estimates, trade-offs, and choices between framed options. Whether extra thinking helps there is a different question, and it is the one this paper asks.

The question has a natural home in dual-process theory. In the human literature, biases such as anchoring and framing are attributed to fast, associative processing, and deliberate reasoning is the mechanism that can override them \cite{tversky1974judgment, evans2013dual}. The analogy between a reasoning trace and this deliberative mode is common in discussions of test-time compute, and it makes a concrete, falsifiable prediction: if the analogy holds, bias should shrink as deliberation grows. To our reading, this dose-response prediction has not been tested under controlled conditions, although its ingredients exist. Benchmarks now document a wide range of cognitive biases in language models \cite{malberg2024comprehensive, echterhoff2024cognitive, binz2023using, knipper2025bias}, and separate work studies how reasoning traces relate to model behavior \cite{turpin2023language, chen2025reasoning}. What is missing is a design that holds the stimuli fixed, pairs reasoning models with matched non-reasoning siblings, and varies deliberation as an experimental factor while checking how much deliberation actually occurred.

The answer matters for practice. If deliberation reduces bias, allocating compute becomes a documented mitigation with a measurable cost-benefit curve. If it does not, then the trust that a longer visible deliberation invites is not backed by better judgment, and bias mitigation has to come from somewhere else. Either outcome changes how a deployed system should be configured; neither can be assumed from capability benchmarks.

We therefore run a dose-response study. From the 30,000-test benchmark of Malberg et al.~\cite{malberg2024comprehensive} we freeze a battery of 30 decision vignettes covering six biases with distinct theoretical roots: anchoring \cite{tversky1974judgment}, the framing effect \cite{tversky1981framing}, loss aversion \cite{kahneman1979prospect}, escalation of commitment \cite{staw1976kneedeep}, the availability heuristic \cite{tversky1973availability}, and confirmation bias \cite{nickerson1998confirmation}. Each vignette exists in a control and a bias-manipulated version, and every response is a choice on a fixed ordinal scale, which yields a bounded, direction-aware bias score. We administer the battery to seven models from four families (Anthropic, OpenAI, DeepSeek, Qwen), crossing each reasoning model with requested thinking ceilings of 0, 1,024, 4,096, and 8,192 tokens and sampling each cell ten times, for 12,350 calls. A third condition on the anchoring items forces the model to restate the anchor before answering. For every call we log the reasoning tokens actually consumed, which turns out to matter: the requested ceiling and the realized deliberation diverge in ways that differ by provider.

The paper makes four contributions:
\begin{itemize}
\item a dose-response design for bias in reasoning models, with matched reasoning/non-reasoning pairs in four model families, a requested thinking ceiling as the manipulated factor, and realized reasoning tokens as the measured dose;
\item evidence that reasoning does not confer resistance: no family shows a reasoning model with lower bias magnitude than its comparator, and no model's bias magnitude reliably declines as realized deliberation grows;
\item a presence-versus-reversal taxonomy in which anchoring alone appears in the human direction while four of the other five biases lean the opposite way in every model tested, with the item-level reliability of each reversal stated;
\item a practical observation: forcing the model to verbalize the anchor lowered anchoring on every anchoring item, which no increase in deliberation did, at the cost of one prompt line.
\end{itemize}
We release the frozen battery, collection harness, scored dataset, per-call token usage, and analysis code.\footnote{\url{https://github.com/obadaKraishan/anchored-minds}}

\section{Related Work}

This section situates the study in three lines of work: measurements of cognitive bias in language models, the test-time reasoning paradigm, and analyses of what reasoning traces actually do.

\subsection{Cognitive biases in language models}
Behavioral experiments borrowed from psychology have become a standard instrument for studying language models. Binz and Schulz probed GPT-3 with classic vignettes and found several human-like errors \cite{binz2023using}. Hagendorff et al.\ reported that intuitive biases present in earlier GPT models weakened or vanished in ChatGPT, an early sign that alignment training reshapes bias profiles \cite{hagendorff2023human}. Itzhak et al.\ showed the complementary effect: instruction tuning can also introduce biases that the base model lacked \cite{itzhak2024instructed}. Larger inventories followed. Echterhoff et al.\ measured biases in managerial decision scenarios \cite{echterhoff2024cognitive}, and Malberg et al.\ built the benchmark used here, 30 biases instantiated in 200 scenarios and validated across 20 models \cite{malberg2024comprehensive}. Knipper et al.\ assessed eight biases across a broad model panel and reported bias-consistent behavior in 18 to 57 percent of instances; among their findings, the reasoning-oriented QwQ-32B and DeepSeek-R1 showed more resistance to framing specifically, with no comparable gain on other biases \cite{knipper2025bias}. Related strands document social forms of bias pressure, including sycophancy \cite{sharma2024towards, fanous2025syceval} and biases that emerge at the population level when model agents interact \cite{ashery2025emergent}. This literature establishes that biases can be measured, but it treats each model as a fixed object; the amount of inference-time deliberation is not manipulated, and whether a model's response moves against the human direction, rather than merely failing to move with it, is not always distinguished.

\subsection{Test-time reasoning}
Chain-of-thought prompting first showed that eliciting intermediate steps improves accuracy \cite{wei2022chain}. Current reasoning models internalize the idea: they are trained to produce an extended thinking phase whose length can be capped by the caller \cite{jaech2024openai, guo2025deepseek, yang2025qwen3}, and allocating more of this test-time compute can beat scaling parameters on verifiable tasks \cite{snell2024scaling, muennighoff2025s1}. The controllable budget is what makes the present study possible: it turns deliberation into an independent variable rather than a model property. As Section~\ref{sec:realized} shows, the control is less direct than the API surface suggests, which is why we measure consumption rather than trusting the request.

\subsection{What reasoning traces do}
A separate line of work cautions against reading the trace as the computation. Turpin et al.\ showed that chain-of-thought explanations can rationalize answers driven by cues the text never mentions \cite{turpin2023language}, and Chen et al.\ found the same pattern inside reasoning models, which verbalize the hints that moved them only a minority of the time \cite{chen2025reasoning}. Shojaee et al.\ documented reasoning effort that collapses precisely where problems get hard \cite{shojaee2025illusion}. These results question the deliberation analogy from the inside, by inspecting traces. We question it from the outside, by testing its behavioral prediction: whatever the trace contains, more of it should produce less bias if it functions as System-2 deliberation. The two approaches are complementary, and our results agree with theirs.

\section{Study Design}

This section describes the research questions, the stimulus battery, the model panel, the experimental conditions and what the gateway did with them, the bias metric and its interpretation, and the collection procedure with its data-quality checks.

\subsection{Research questions}
The study is organized around four questions, fixed before data collection.
\begin{itemize}
\item \textbf{RQ1.} Do reasoning models show smaller bias magnitudes than matched non-reasoning models of the same family?
\item \textbf{RQ2.} Does bias decrease as deliberation grows?
\item \textbf{RQ3.} Which of the six biases are present in the human direction, absent, or reversed, and which respond to deliberation?
\item \textbf{RQ4.} Does forcing the model to verbalize the anchor change anchoring susceptibility?
\end{itemize}

\subsection{Benchmark and battery}
Stimuli come from the cognitive-bias benchmark of Malberg et al.~\cite{malberg2024comprehensive}, which instantiates each bias as paired prompts: a control version of a managerial decision scenario and a treatment version containing the bias manipulation, both answered on a fixed ordinal scale presented as numbered options. We selected six biases on theoretical grounds, covering numeric estimation (anchoring), choice architecture (framing, loss aversion, escalation of commitment), and belief-driven judgment (availability, confirmation). For each bias we sampled five items from distinct scenarios with a fixed seed (42) and froze the resulting 30-item battery before any experimental call was made; the battery did not change afterwards. Ten items use a 7-point scale and twenty use an 11-point scale. Each benchmark item carries a direction parameter $k \in \{-1, +1\}$ and, where applicable, a scale-reversal flag for the treatment prompt, which we apply as specified so that positive scores always mean movement in the direction the bias predicts for humans.

\subsection{Model panel}
Table~\ref{tab:design} lists the panel: seven models from four families, chosen to span capability tiers and open versus closed weights while keeping a matched reasoning/non-reasoning pair inside each family. All calls were routed through a single API gateway (OpenRouter) so that the reasoning budget could be requested with one uniform parameter. Two properties of the panel required design decisions. For the OpenAI family no separate non-reasoning sibling of GPT-5.1 was available, so the comparator is the same model at budget~0 with thinking disabled, which holds the weights constant and toggles only deliberation. DeepSeek-R1 and the Qwen3 thinking variant cannot disable thinking at all: requests to turn it off were rejected or ignored, and at a zero budget the models thought anyway. We therefore treat budget~0 as undefined for these two models and use their instruct siblings (DeepSeek-V3, Qwen3-Instruct) as the family zero points, which is what the paired design was built for. This means the reasoning contrast is held at constant weights for the Claude and OpenAI families and crosses a sibling boundary for DeepSeek and Qwen; we return to this in the limitations.

\begin{table}[!t]
\centering
\caption{Model panel and design cells per model. Budgets are requested extended-thinking token ceilings.}
\label{tab:design}
\footnotesize
\setlength{\tabcolsep}{3.5pt}
\begin{tabular}{lllll}
\hline
Model & Family & Type & Budgets & Cells \\
\hline
claude-haiku-4.5-thinking & claude & reasoning & 0, 1k, 4k, 8k & 120 \\
claude-haiku-4.5 & claude & instruct & 0 & 30 \\
gpt-5.1-reasoning & openai & reasoning & 0, 1k, 4k, 8k & 120 \\
deepseek-r1 & deepseek & no-disable & 1k, 4k, 8k & 90 \\
deepseek-v3 & deepseek & instruct & 0 & 30 \\
qwen3-thinking & qwen & no-disable & 1k, 4k, 8k & 90 \\
qwen3-instruct & qwen & instruct & 0 & 30 \\
\hline
\end{tabular}
\par\smallskip\footnotesize ``no-disable'' models cannot switch thinking off; their instruct siblings provide the family's budget-0 point.
\end{table}

\subsection{Conditions and budget manipulation}
Every battery item was administered in a control and a treatment condition. Anchoring items received a third condition for RQ4: the treatment prompt plus one instruction to restate every number mentioned in the prompt before answering. All prompts ended with a fixed answer-format instruction (``Answer: Option $\langle$number$\rangle$''). Reasoning models ran at requested ceilings of 0, 1,024, 4,096, and 8,192 thinking tokens (the two no-disable models at the three positive ceilings); instruct models ran at budget 0 only. Each model $\times$ budget $\times$ item $\times$ condition cell was sampled ten times at temperature 1.0, giving 12,350 calls in total. Requests used asynchronous collection with retry and checkpointing; the token ceiling for thinking calls reserved a fixed answer allowance on top of the budget, which a pilot showed was necessary to keep long deliberations from truncating the visible answer.

\subsection{Requested ceiling versus realized deliberation}
\label{sec:realized}
A requested ceiling is an instruction to the provider, not a measurement of what the model did. The gateway's documentation states that Anthropic models honour the ceiling directly, that OpenAI reasoning models accept only an effort level and have the ceiling converted to one, and says nothing about how open-weight providers treat it. We therefore logged the reasoning tokens reported in the usage field of every response and treat that count, rather than the request, as the dose. Table~\ref{tab:tokens} summarises the result, and it differs by provider in three ways.

For Claude, the ceiling was honoured and consumption rose with it, from zero at budget~0 to a mean of 437, 575, and 620 tokens at the three positive ceilings; no call exceeded its ceiling, and the model never came close to the larger ones. For GPT-5.1, thinking was off at budget~0 and nearly constant above it, at 116 to 118 tokens regardless of the requested ceiling, which is the effort-tier conversion at work: the three requested levels produced one realized level. For DeepSeek-R1 and Qwen3, the ceiling was not enforced. Both models consumed between 1,074 and 1,617 tokens on average at every setting including budget~0, and at the 1,024 ceiling 79\% of R1 calls and 52\% of Qwen3 calls exceeded it. Consumption rose slightly between the 1,024 and 4,096 settings and was flat beyond.

Two consequences follow for the analysis. The budget-0 contrast for Claude and GPT-5.1 is a true zero-deliberation comparison at constant weights. And the dose-response question (RQ2) is answered on realized tokens: per model, we regress bias on $\log_2(\text{realized tokens}+1)$ within the range that model actually produced, rather than on the requested label.

\begin{table}[!t]
\centering
\caption{Realized reasoning tokens per call (cell means) against the requested ceiling.}
\label{tab:tokens}
\footnotesize
\setlength{\tabcolsep}{4pt}
\begin{tabular}{lrrrr}
\hline
Model & $b=0$ & $b=1{,}024$ & $b=4{,}096$ & $b=8{,}192$ \\
\hline
claude-haiku-4.5-thinking & 0 & 437 & 575 & 620 \\
gpt-5.1-reasoning & 0 & 116 & 118 & 118 \\
deepseek-r1 & 1,188 & 1,555 & 1,605 & 1,617 \\
qwen3-thinking & 1,074 & 1,321 & 1,440 & 1,436 \\
\hline
\end{tabular}
\par\smallskip\footnotesize Honoured for Claude; converted to an effort tier for GPT-5.1 (flat consumption); not enforced for DeepSeek-R1 and Qwen3, where 79\% and 52\% of calls exceeded the 1,024 ceiling. The $b=0$ cells for the two no-disable models are the excluded conditions described in the text.
\end{table}

\subsection{Bias metric}
Let $\mu_C$ and $\mu_T$ be the mean chosen option index across the ten samples of a cell's control and treatment condition, on a scale with $n$ options, after applying the benchmark's treatment-scale reversal where flagged. The cell's bias score is
\begin{equation}
s \;=\; k \cdot \frac{\mu_T - \mu_C}{\,n - 1\,} \;\in\; [-1,\, +1],
\label{eq:score}
\end{equation}
where $k$ is the benchmark's direction parameter. A positive $s$ means the treatment moved the model in the direction the bias predicts for humans; a negative $s$ means it moved the model the opposite way. We analyze $s$ (signed) and $|s|$ (magnitude, i.e., susceptibility regardless of direction). The verbalized condition yields an analogous score $s_v$ against the same control.

What the score measures deserves one paragraph. The control and treatment versions of an item carry the same decision-relevant information; by construction they differ only in the manipulation (a numeric anchor, a gain or loss frame, a prior commitment, and so on) that the bias literature identifies as normatively irrelevant to the choice. Any systematic shift between the two therefore indexes sensitivity to an irrelevant feature of the prompt. That is the sense in which we use ``bias'': susceptibility to the manipulation, not a judgment about the quality of either answer in isolation. Magnitude $|s|$ is the size of that susceptibility, and the sign says whether the shift runs with or against the human pattern. A model with $|s| = 0$ is one whose answer does not depend on the irrelevant feature; a model with large $|s|$ in either direction is one whose answer does.

\subsection{Procedure and data quality}
All 12,350 calls completed without API failures. Answers were extracted with a tiered, range-aware parser (strict format first, then boxed and free-text patterns, all restricted to the item's option range so that echoed stimulus numbers cannot be mistaken for answers). On analyzable data the parse rate was 99.4\%, at or near 100\% for five of the seven models. Two exclusions are documented rather than repaired. First, the budget-0 responses of the two no-disable models are excluded by design as described above. Second, Qwen3-Instruct provided a raw quantity (for example, a percentage) instead of an option index in 7.2\% of its responses, mostly on anchoring estimation items; these responses were excluded, leaving seven of its 65 condition cells below eight samples but every cell with at least two; five scored cells are affected, 1.0\% of the 510-cell grid. The scored dataset contains one row per model $\times$ budget $\times$ item cell (510 rows).

The unit of independent observation is the item. Where a comparison pools across models or budgets, the same 30 items recur, so we aggregate to the item level before testing and report item-clustered bootstrap intervals. Where a test is confined to one bias, only its five items are independent, so per-bias tests use the $t$ distribution with four degrees of freedom on the five item means; Wald tests from mixed models assume many clusters and overstate significance at this size. Mixed models with a random item intercept are used where all 30 items enter. Within-family contrasts use paired $t$-tests and Wilcoxon signed-rank tests over the 30 items with Cohen's $d_z$; dose-response uses per-item ordinary-least-squares slopes on $\log_2(\text{realized tokens}+1)$ together with mixed models; Holm correction is applied within each family of tests.

\section{Results}

This section reports the four research questions in turn. Descriptively, mean signed scores are negative for every model at every budget (range $-0.23$ to $-0.10$), which already hints that the dominant behavior is movement against the human bias directions; the exception, anchoring, is isolated in Sec.~\ref{sec:rq3}.

\subsection{RQ1: Reasoning versus matched non-reasoning models}
Table~\ref{tab:rq1} compares bias magnitude $|s|$ for each reasoning model (averaged over its positive budgets) against its family comparator at budget 0, paired by item. In no family is the reasoning model less biased. All four point estimates lean the same way, with the reasoning model slightly higher ($d_z = 0.08$ to $0.26$), and none is reliable after Holm correction. Pooled across families by averaging each item's four differences, so that the 30 items remain the independent units, the difference is $\Delta = +0.031$, 95\% bootstrap CI $[-0.010, +0.074]$, $t(29) = 1.45$, $p = .157$, $d_z = 0.27$; the Wilcoxon test agrees ($p = .253$), as does a mixed model with a random item intercept ($z = 1.45$, $p = .147$). We read this as no evidence that reasoning models are the lower-bias choice on these tasks. The consistent direction across families is worth noting but not worth a claim; the pooled interval includes zero. On signed scores no family shows a reliable difference either.

\begin{table}[!t]
\centering
\caption{RQ1: bias magnitude $|s|$ for reasoning models (budgets $>0$) versus matched non-reasoning comparators, paired by item ($n=30$).}
\label{tab:rq1}
\footnotesize
\setlength{\tabcolsep}{2.6pt}
\begin{tabular}{lccccccc}
\hline
Family & $M_{\text{reas}}$ & $M_{\text{instr}}$ & $\Delta$ & 95\% CI & $t(29)$ & $p_{\text{Holm}}$ & $d_z$ \\
\hline
claude   & 0.271 & 0.244 & $+0.027$ & $[-0.024, 0.088]$ & 0.95 & .714 & 0.17 \\
deepseek & 0.275 & 0.249 & $+0.026$ & $[-0.013, 0.070]$ & 1.20 & .714 & 0.22 \\
openai   & 0.299 & 0.243 & $+0.056$ & $[-0.014, 0.140]$ & 1.40 & .686 & 0.26 \\
qwen     & 0.301 & 0.287 & $+0.015$ & $[-0.046, 0.079]$ & 0.45 & .714 & 0.08 \\
\hline
\end{tabular}
\par\smallskip\footnotesize Positive $\Delta$ = reasoning model more biased. Pooled at the item level: $\Delta = +0.031$, $t(29)=1.45$, $p=.157$, $d_z=0.27$.
\end{table}

\subsection{RQ2: Dose-response over realized deliberation}
\begin{figure}[!t]
\centering
\includegraphics[width=\columnwidth]{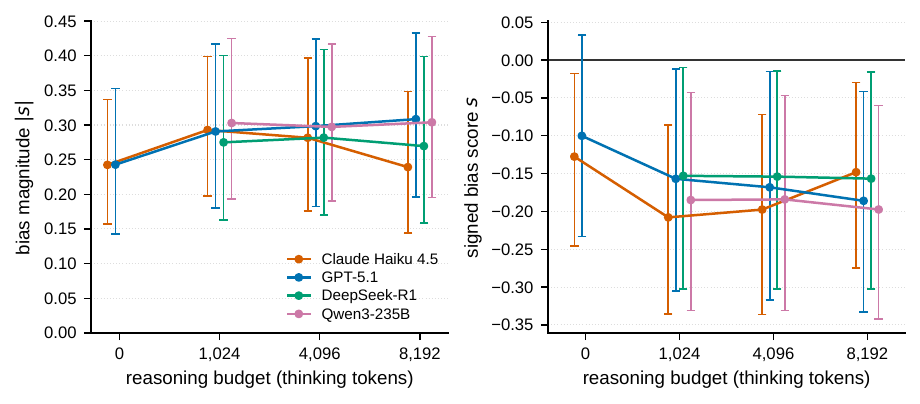}
\caption{Bias against the requested reasoning ceiling (error bars: 95\% bootstrap CIs). Left: magnitude $|s|$. Right: signed score $s$. The $x$-axis is the requested ceiling; Table~\ref{tab:tokens} gives the deliberation each setting actually produced, and Table~\ref{tab:rq2} tests the trend on that realized quantity.}
\label{fig:dose}
\end{figure}

Fig.~\ref{fig:dose} plots bias against the requested ceiling for orientation; Table~\ref{tab:rq2} gives the statistics that matter, computed on realized tokens within each model's observed range. No model shows a reliable decline in magnitude. For Claude, whose realized deliberation spans 0 to about 800 tokens per cell, the per-item slope of $|s|$ is indistinguishable from zero ($+0.0029$ per doubling, $t(29) = 0.91$, $p = .372$; mixed model $p = .259$). For DeepSeek-R1 (roughly 700 to 2,800 tokens per cell) and Qwen3 (roughly 500 to 3,400) the slopes are likewise null ($p = .958$ and $p = .213$ in the mixed models). GPT-5.1 is the one model whose magnitude rises with realized tokens ($\beta = +0.0084$, $p = .018$), but its realized range is 0 to about 225 tokens, so this is in effect the budget-0 versus any-thinking contrast of RQ1 rather than a dose gradient.

Where deliberation does anything, it is to the signed score, and it moves away from the human direction rather than toward zero: Claude ($\beta = -0.0059$ per doubling, $p = .018$) and GPT-5.1 ($\beta = -0.0104$, $p = .010$) both drift further negative as realized tokens increase; DeepSeek-R1 and Qwen3 do not ($p = .776$ and $p = .075$, the latter in the positive direction). Taken together, across the range of deliberation these models actually produced, which the requested ceilings expanded for Claude but not for the others, no model's susceptibility to the manipulations reliably decreased.

\begin{table}[!t]
\centering
\caption{RQ2: trend of bias on $\log_2(\text{realized reasoning tokens}+1)$, per model, within the realized range.}
\label{tab:rq2}
\footnotesize
\setlength{\tabcolsep}{3pt}
\begin{tabular}{lcccc}
\hline
Model & Range (tokens) & slope $t(29)$, $p$ & $\beta_{|s|}$ ($p$) & $\beta_{s}$ ($p$) \\
\hline
claude-th.  & 0--805    & $0.91$, .372  & $+0.003$ (.259) & $-0.006$ (.018) \\
gpt-5.1     & 0--225    & $1.36$, .184  & $+0.008$ (.018) & $-0.010$ (.010) \\
deepseek-r1 & 707--2807 & $-1.63$, .114 & $+0.003$ (.958) & $-0.018$ (.776) \\
qwen3-th.   & 534--3440 & $0.25$, .805  & $-0.066$ (.213) & $+0.093$ (.075) \\
\hline
\end{tabular}
\par\smallskip\footnotesize Range is the span of cell-mean realized tokens. ``slope'' tests the per-item OLS slope of $|s|$ against zero; $\beta$ coefficients are from mixed models with random item intercepts, per doubling of realized tokens. No $|s|$ trend is significantly negative.
\end{table}

\subsection{RQ3: Which biases exist, and in which direction}
\label{sec:rq3}
\begin{figure*}[!t]
\centering
\includegraphics[width=0.86\textwidth]{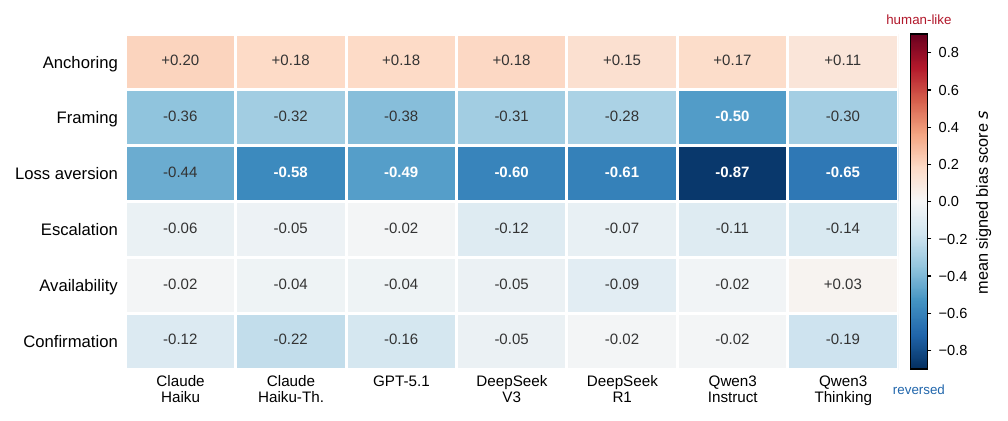}
\caption{Mean signed bias score by bias type and model. Red indicates movement in the human bias direction, blue the reverse. Anchoring is the only red row. The other biases are blue in every model, reasoning and instruct alike, with one exception: availability in Qwen3-thinking.}
\label{fig:heatmap}
\end{figure*}

Table~\ref{tab:rq3} classifies each bias from the signed score over all 85 model $\times$ budget cells per bias, testing the five item means against zero with $t(4)$ so that inference respects the five items that instantiate each bias; Fig.~\ref{fig:heatmap} shows the full bias $\times$ model map. Two statements have different evidential status here, and we keep them apart.

The first is about models. In all seven models, anchoring is positive in mean and framing, escalation of commitment, confirmation bias, and loss aversion are negative (Fig.~\ref{fig:heatmap}); availability is negative in six of the seven. The pattern replicates across four training pipelines and both open and closed weights. Anchoring is present in the human direction and strongly so ($M = +0.163$, 95\% CI $[+0.111, +0.215]$, $t(4) = 5.02$, $p_{\text{Holm}} = .037$, $d = 1.89$): a numeric anchor pulls estimates toward itself in every model, from $+0.11$ (Qwen3-thinking) to $+0.20$ (Claude instruct).

The second is about biases as constructs, which requires generalizing over items, and with five items per bias that test is weaker. The framing effect, which in people flips risk preferences between gain and loss frames \cite{tversky1981framing}, runs backwards here and reliably so ($M = -0.336$, $t(4) = -7.57$, $p_{\text{Holm}} = .010$, $d = -2.35$). Escalation of commitment is reversed in every model and its interval excludes zero ($M = -0.070$, $[-0.119, -0.031]$), but its item-level test does not survive correction ($t(4) = -2.79$, $p_{\text{Holm}} = .197$); confirmation bias is similar ($M = -0.139$, $p_{\text{Holm}} = .227$). Loss aversion shows the largest mean reversal in the battery ($M = -0.586$, reaching $-0.87$ for Qwen3-Instruct) but also the widest item-clustered interval ($[-0.914, +0.031]$, $p_{\text{Holm}} = .256$): the five loss-aversion items disagree about how far the models lean, even though all seven models lean the same way. Availability is small, not reliable at the item level ($M = -0.035$, $p_{\text{Holm}} = .256$), and positive in one model (Qwen3-thinking, $+0.03$), so we classify it as absent. We therefore describe framing as reversed; escalation, confirmation, and loss aversion as directional, meaning opposite to the human direction in every model with item-level reliability that five items cannot establish; and availability as absent. No bias qualifies as deliberation-responsive; all six per-bias slope tests on realized tokens are null after correction.

\begin{table}[!t]
\centering
\caption{RQ3: presence classification per bias. Signed score vs.\ 0 over 85 cells; $t(4)$ on the five item means; item-clustered bootstrap CIs; Holm-corrected.}
\label{tab:rq3}
\footnotesize
\setlength{\tabcolsep}{1.9pt}
\begin{tabular}{lcccccl}
\hline
Bias & $M$ & 95\% CI & $t(4)$ & $p_{\text{H}}$ & $d$ & Class \\
\hline
Anchoring    & $+.163$ & $[+.111, +.215]$ & $5.02$  & .037 & $+1.89$ & Present \\
Framing      & $-.336$ & $[-.403, -.245]$ & $-7.57$ & .010 & $-2.35$ & Reversed \\
Escalation   & $-.070$ & $[-.119, -.031]$ & $-2.79$ & .197 & $-0.68$ & Directional \\
Confirmation & $-.139$ & $[-.249, -.043]$ & $-2.38$ & .227 & $-0.65$ & Directional \\
Loss aversion& $-.586$ & $[-.914, +.031]$ & $-1.91$ & .256 & $-0.88$ & Directional \\
Availability & $-.035$ & $[-.068, +.005]$ & $-1.71$ & .256 & $-0.42$ & Absent \\
\hline
\end{tabular}
\par\smallskip\footnotesize $d$ is the cell-level standardized mean. ``Present'' = human-like direction; ``Reversed'' = reliable shift against it at the item level; ``Directional'' = opposite to the human direction in all seven models, but the item-level test does not reach .05 after correction with five items per bias; ``Absent'' = neither reliable nor consistent in sign across models.
\end{table}

\subsection{RQ4: Forcing the model to verbalize the anchor}
\begin{figure}[!t]
\centering
\includegraphics[width=\columnwidth]{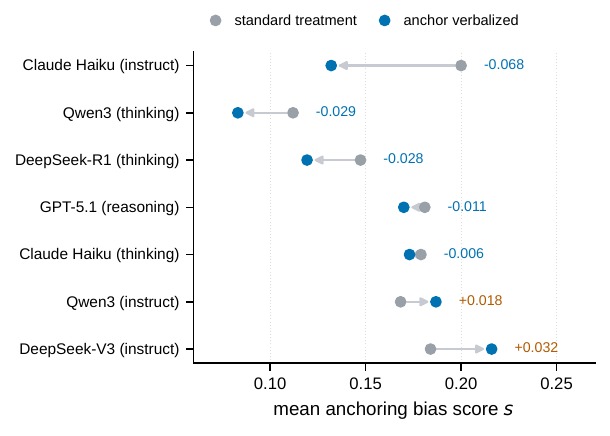}
\caption{Anchoring with and without forced anchor verbalization, by model. Verbalization lowers anchoring in five of seven models and on all five anchoring items; at the item level the reduction falls short of significance ($t(4)=-2.65$, $p=.057$).}
\label{fig:verb}
\end{figure}

The verbalized condition adds one instruction to the anchoring treatment: restate every number in the prompt before answering. The change runs in the helpful direction on every one of the five anchoring items, with item means from $-0.004$ to $-0.036$ and a pooled mean of $\Delta = -0.015$ (95\% CI $[-0.026, -0.004]$ over cells). Tested at the item level, with five independent items, the effect falls just short of significance ($t(4) = -2.65$, $p = .057$; a sign test on five of five items gives $p = .063$). Five of the seven models also move in that direction (Fig.~\ref{fig:verb}). We therefore report the verbalization effect as directional rather than established: consistent across items and most models, small, and resting on five items. Its comparison point is still informative. A single prompt line moved anchoring the helpful way on every item, while no level of realized deliberation in RQ2 moved it at all.

\section{Discussion}

This section interprets the main findings and draws out what they mean for deployment and for the deliberation analogy.

\subsection{A bias landscape that leans the wrong way}
The most consistent result is not about reasoning at all. Across seven models spanning four training pipelines and both open and closed weights, anchoring runs in the human direction and four of the other five biases lean against it in every model (Fig.~\ref{fig:heatmap}), while availability is small and inconsistent. At the item level that reversal is firmly established for framing and directional for escalation, confirmation, and loss aversion, so the strong form of the claim rests on one bias and the weaker form on four.

This picture departs from the benchmark results of Malberg et al.~\cite{malberg2024comprehensive}, who report bias-consistent behavior across their model panel, and from the one-sided framing in Knipper et al.~\cite{knipper2025bias}, where reasoning models show more resistance to framing. Three differences in method bear on the comparison. We sample ten responses per cell at temperature 1.0 and score the mean shift, rather than scoring single responses; we use five items per bias drawn from their pools; and our panel consists of models released in 2025 and later, after the alignment generations those studies evaluated. We cannot attribute the divergence to any one of these from our data. What we can say is that, for four biases, the sign of the reversal is uniform across our seven models, which is not what item sampling alone would produce, and that a one-sided ``resistance'' finding and a signed ``reversal'' finding are compatible: both say that reasoning models do not frame like people, and the signed metric adds that they move the other way.

One reading of the reversal is that this is what heavy preference tuning looks like from the outside. Hagendorff et al.\ observed human-like biases weakening from GPT-3 to ChatGPT \cite{hagendorff2023human}, and Itzhak et al.\ showed that instruction tuning reshapes bias profiles in both directions \cite{itzhak2024instructed}; our data are consistent with that trajectory continuing past neutrality. Loss aversion is the suggestive case: its reversal is what a risk-neutral expected-value maximizer would produce on these choice pairs, and models trained to give ``rational'' answers may have internalized that norm, though with five items and a wide interval we present this as a hypothesis rather than a finding. Whatever the mechanism, the practical upshot holds for the biases that are reliably reversed: the familiar human bias catalog is the wrong checklist for auditing these systems. A review that asks ``is the model framing-sensitive like a person?'' will answer no and miss that the model deviates from frame-invariance just as far the other way.

\subsection{No debiasing dividend from deliberation}
Against this backdrop, the reasoning results point one way. Matched within families, reasoning models are not less biased than their comparators (RQ1); the point estimates lean toward slightly more bias in every family, but the pooled interval includes zero and we do not claim a difference. Across the deliberation these models actually produced (RQ2), no model's bias magnitude reliably declined. The qualification that the realized-token analysis adds is important: for GPT-5.1 the ceiling did not change how much the model thought, and for DeepSeek-R1 and Qwen3 the ceiling was not enforced, so those three models provide a narrow dose range. Claude is the one model whose deliberation the ceiling controlled, from zero to several hundred tokens, and across that range its bias magnitude was flat while its signed score drifted slightly further from the human direction. The dual-process prediction, more deliberation, less bias, finds no support in the range tested. This does not mean the thinking is idle; on verifiable tasks it demonstrably helps \cite{snell2024scaling}. It means the deliberation does not function as a bias corrector here, which fits the trace-level evidence that reasoning text and answer-driving computation can come apart \cite{turpin2023language, chen2025reasoning, shojaee2025illusion}. Anchoring makes the point concretely: the anchor keeps its pull at the highest realized deliberation we observed, so whatever those tokens do, they do not include discounting the number that is biasing the answer.

A methodological lesson sits beside the substantive one. A requested thinking ceiling, sent through a multi-provider gateway, produced three different realities: a ceiling that was honoured but rarely approached, a ceiling converted into a single effort setting, and a ceiling that was ignored. Studies that manipulate reasoning budget and do not log consumption cannot tell these cases apart. We recommend reporting realized reasoning tokens per condition as a matter of course.

\subsection{Anchor restatement versus deliberation}
The one intervention that moved anchoring was not more compute but a redirection of attention: asking the model to restate the anchor before answering lowered anchoring on all five items (RQ4). The size is modest, one model (DeepSeek-V3) shifted the other way, and with five items the effect does not reach significance, so we present this as a cheap mitigation worth testing rather than an established one. Still, the contrast with RQ2 is the useful part. For this bias, one line of prompt engineering did what the available deliberation did not, which suggests that targeted prompting and per-bias audits are currently a better use of effort than budget increases when the goal is judgment quality rather than task accuracy.

\subsection{Implications for practice}
Three recommendations follow. First, do not assume the reasoning tier of a model family is the lower-bias option; in our panel it never was, and the evidence gives no reason to expect it. Second, audit deployed models per bias and per direction, since for four of the six biases the sign of the deviation was model-agnostic here but its size varied by family. Third, where numeric anchors are a known hazard, an anchor-restating instruction is a low-cost mitigation worth testing in context.

\section{Limitations}

The design has boundaries worth stating plainly. The stimuli are vignettes from a single benchmark family; vignette-based measurement is the standard paradigm in both the human and machine literatures \cite{tversky1974judgment, malberg2024comprehensive}, but effects can be sensitive to phrasing, and five items per bias is a narrow base. That narrowness is why three of the four reversals and the RQ4 verbalization effect are reported as directional rather than reliable, and why availability is classified as absent: with five items per bias, an item-level test has four degrees of freedom, and effects that are consistent in sign across every item and model can still fall short of significance. The deliberation manipulation did not work uniformly: realized reasoning tokens rose with the requested ceiling for Claude, were constant for GPT-5.1, and were uncapped for DeepSeek-R1 and Qwen3, so the dose-response conclusion is bounded by the realized ranges in Table~\ref{tab:rq2}, and we cannot speak to deliberation beyond roughly 3,400 tokens per cell. For the two no-disable models the reasoning contrast crosses a sibling boundary and so confounds deliberation with weights; the Claude and OpenAI contrasts hold weights constant and are the cleaner evidence for RQ1. Our claims are behavioral: the study measures what the models do, not why, and the overcorrection reading of the reversals is an interpretation the data are consistent with, not a demonstrated mechanism. Finally, all calls went through one routing provider during one collection window; model updates behind fixed identifiers are a general reproducibility hazard, which the released harness, frozen battery, and per-call usage logs mitigate but cannot remove.

\section{Conclusion}

We asked whether test-time reasoning immunizes language models against human cognitive biases, and within the deliberation these models actually produced, the answer is no on every axis measured: reasoning models are not less biased than matched siblings, bias magnitude does not reliably fall as realized deliberation grows, and the one bias that behaves like its human counterpart, anchoring, is as strong at the highest deliberation observed as at none. What the data show instead is a bias landscape that leans against the human catalog, consistently across seven models for four biases and reliably so for framing, together with a small, cheap prompt intervention that moved anchoring on every item where compute did not. A requested thinking budget turned out to be a poor proxy for thinking done, which is itself a finding this literature should absorb. For a venue concerned with cognitive machine intelligence, the message cuts both ways. The dual-process analogy earns its keep as a hypothesis generator, but its central prediction fails in the range tested, and evaluation practice should follow the data: measure biases in both directions, per model, log the deliberation that actually occurred, and treat requested budget as a cost knob rather than a rationality guarantee.

\bibliographystyle{IEEEtran}
\bibliography{references}

\end{document}